\documentclass[letterpaper, 10 pt, conference]{ieeeconf}  

\IEEEoverridecommandlockouts                              

\let\labelindent\relax

\usepackage{subcaption}

\usepackage{graphics} 
\usepackage{graphicx}
\usepackage{graphicx}
\usepackage{comment}
\usepackage{amsfonts}
\usepackage{amsmath} 
\usepackage{amssymb}  
\usepackage{enumitem}
\usepackage{float}
\usepackage{algorithm} 
\usepackage[noend]{algpseudocode}
\usepackage{multicol}
\usepackage{xcolor}
\usepackage{fullwidth}
\usepackage{url}
\usepackage{array}
\newcolumntype{P}[1]{>{\centering\arraybackslash}p{#1}}
\usepackage[T1]{fontenc}

\definecolor{Concrete}{RGB}{80,90,90}
\definecolor{Ground}{RGB}{100,100,50}
\definecolor{Vegetation}{RGB}{17,150,2}
\definecolor{Water}{RGB}{20,20,200}
\definecolor{Wood}{RGB}{250,45,8}
\definecolor{Tarp}{RGB}{255,255,5}

\title{\LARGE \bf
Real-Time Semantic Segmentation using Hyperspectral Images for Mapping Unstructured and Unknown Environments 
}

\author{ Anthony Medellin$^{1*}$, Anant Bhamri$^{1*}$, Reza Langari$^{1}$, and Swaminathan Gopalswamy$^{1}$
\thanks{$^{*}$ Equal contribution \newline $^{1}$ Anthony Medellin, Anant Bhamri, Reza Langari, and Swaminathan Gopalswamy are with Texas A\&M University, College Station, TX, USA      {\tt\small \{ antmedellin, anantbhamri, rlangari, sgopalswamy\}@tamu.edu}}%
}

\begin{document}

\maketitle
\thispagestyle{empty}
\pagestyle{empty}

\begin{abstract}

Autonomous navigation in unstructured off-road environments is greatly improved by semantic scene understanding. Conventional image processing algorithms are difficult to implement and lack robustness due to a lack of structure and high variability across off-road environments. The use of neural networks and machine learning can overcome the previous challenges but they require large labeled data sets for training. In our work we propose the use of hyperspectral images for real-time pixel-wise semantic classification and segmentation, without the need of any prior training data. The resulting segmented image is processed to extract, filter, and approximate objects as polygons, using a polygon approximation algorithm. The resulting polygons are then used to generate a semantic map of the environment. Using our framework. we show the capability to add new semantic classes in run-time for classification. The proposed methodology is also shown to operate in real-time and produce outputs at a frequency of $1$Hz, using high resolution hyperspectral images. 

\end{abstract}

\section{Introduction}

Enabling autonomy in various domains and environments has been gathering a lot of interest recently. The known and predictable structure of urban terrains makes it easy to extract the underlying semantic information but the same techniques cannot be applied to off-road terrains. In complex, unstructured settings the lack of semantic information like the terrain type and traversability, makes enabling autonomy an ongoing research challenge. The ability to classify and understand the complex scene in which an autonomous agent needs to operate will help make them a reality in a wider range of applications including defense, agriculture, conservation, and search and rescue. 

The current state of the art techniques \cite{ jiang2021lidarnet, Huang_2019_ICCV}, use either a camera or a LiDAR for this semantic scene understanding, but both approaches rely on machine learning algorithms or neural networks for this task. The inherent disadvantage with this approach is two fold. Firstly, it requires a large training data set in which the various semantic classes should be labeled and second, if the algorithm comes across a class which was absent from the training data set, there is no way to easily add more classes during run-time. Further, even if we want to add this new class later, it requires generation of a larger labeled training data set \cite{cordts2016cityscapes, mo2022review}.

\begin{figure}
    \centering
    \includegraphics[width=3in]{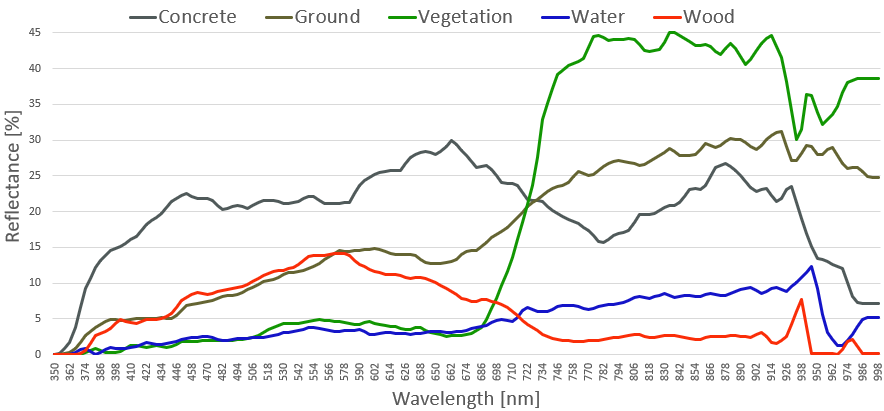}
    \caption{{Example spectral data for different materials as seen from a hyperspectral camera }}
    \vspace{-1.8em}
    \label{fig:SpectralShape}
\end{figure}

To solve the above mentioned problems we suggest the use of hyperspectral cameras. These cameras capture information across multiple bands in the electromagnetic spectrum, unlike a standard RGB camera which only captures information from three bands (Red, Blue and Green) in the visible region of the electromagnetic spectrum. Fig. \ref{fig:SpectralShape} shows the spectral data for 5 different materials (used in the experiments), as captured from a hyperspectral camera. It shows the reflectance value of the different classes with respect to wavelength. The presence of these additional bands allows us to classify the pixels at the material level \cite{imani2020overview}. Thus, we need only one pixel labeled as a reference for classifying a material or object. 

In this paper, we describe a novel framework to directly generate inputs for a 2D semantic map, representing the environment with both geometrical and semantic properties, using only hyperspectral images. Further, we provide a way to label unknown semantic classes in real-time using the spectral properties of each material. The segmented images are then processed to extract polygonal features which are used to initialize the semantic map. Moreover, we demonstrate the capability to filter and approximate polygonal features based on their size. This is helpful if we need to reduce noise or share a lower fidelity map between multiple agents due to bandwidth or memory constraints.

The main contribution of our work is to perform semantic scene understanding with or without the use of any prior training data and the ability to add new semantic classes to the reference database in real-time. We show the effectiveness of our approach by generating a semantic map of an unknown environment in real-time, without any reference database to start with. For ease of application, we can also start with a predefined database \cite{meerdink2019ecostress} which can be expanded if a new semantic class is observed. 

\section{Related Work}

The goal of semantic segmentation is to classify data points from an exteroceptive sensor into a predefined category. Researchers have been using cameras, LiDARs, RaDARs, or their combination for semantic segmentation and scene understanding \cite{feng2020deep}. Before the widespread use of deep learning techniques, most image-based methods focused on using algorithms based on edge detection, boundary tracing, and region growing \cite{sam2012survey, mivcuvslik2009semantic, dollar2006supervised} and point cloud segmentation was explored using geometric constraints and assumed prior knowledge \cite{zhang2019review, Xie_2020}. These techniques have limitations in handling unstructured complex environments. Neural network based algorithms are used to overcome this challenge, but they require the algorithm to be trained on large, labeled data sets, a task that is both time consuming and resource intensive \cite{mo2022review, maturana2018real}. Moreover, the performance degrades if one is operating in conditions which are not captured by the training data set.

Hyperspectral cameras capture images with multiple spectral bands, which help in creating a unique spectral shape for the various materials in the image. This allows reduction of the training data required for semantic classification. 
Classification and segmentation of hyperspectral images use Convolutional Neural Networks (CNNs), clustering, or nearest-neighbor to extract spatial and spectral features \cite{lee2017going, li2015skin, jakubczyk2022hyperspectral}. Also, \cite{kruse1993spectral} shows how the existing spectral database can be used for classification, but the use of large number of classes in the database increases the computation time of the spectral similarity algorithms. Our work provides a framework which can be used with an existing reference database or one can be created by the user in run-time based on the specific task or mission requirements. \cite{liyanage2020hyperspectral} uses hyperspectral images to annotate RGB images to train a CNN for semantic segmentation of unstructured terrains and \cite{jakubczyk2022hyperspectral} generates the spectral data for multiple semantic classes by labeling multiple pixels from each class. This makes the addition of new semantic classes cumbersome. To address this issue, we provide a methodology to add new semantic classes during the mission in real time.

Availability of the semantically segmented information facilitates planning and decision making. For example, \cite{jakubczyk2022hyperspectral, maturana2018real} generate occupancy grid based maps using a LiDAR and the segmented images used for path planning. The use of knowledge frameworks to represent the underlying implicit information that is present in the environment is explored in \cite{hanheide2017robot, tenorth2009knowrob} which can assist the agent in higher level plans for completing a task but they have to rely on other sources for generating the finer inputs for controls. Our proposed approach can readily combine the semantic classification and the abstracted geometric information of the features (as polygons) to create an ontological representation that could then lead to automated reasoning and decision making. 

\section{Methodology}

\begin{figure}
    \centering
    \includegraphics[width=3.4in]{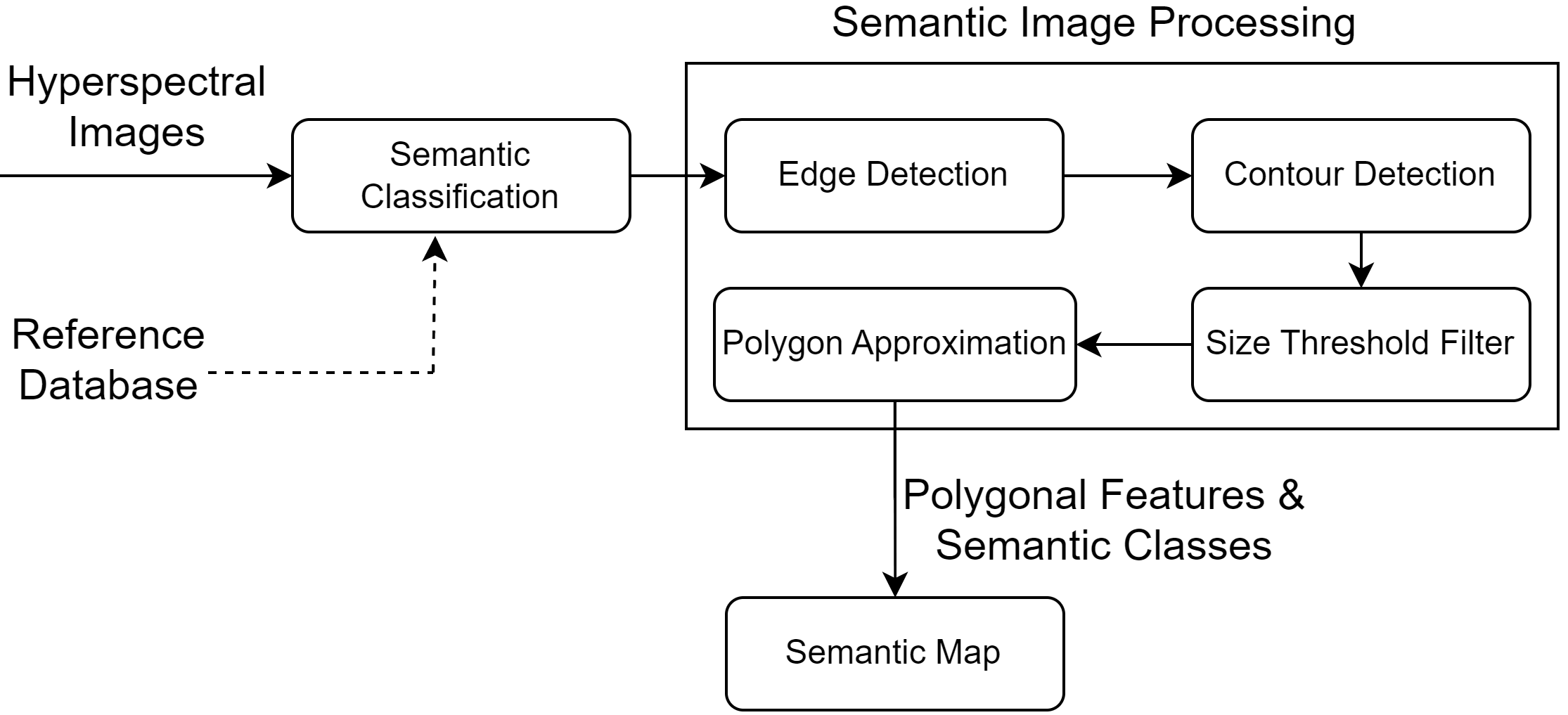}
    \caption{ {Architecture of the Proposed Semantic Mapping Framework}}
    \label{fig:Architechture}
    \vspace{-1.8em}
\end{figure}

\subsection{Hyperspectral Image Classification}
Our assumption is that no prior information is known about the environment before an image is captured. A spectral database will be created during run-time by a user, and will contain the reference spectra of the user-defined semantic classes. A reference database can be generated by the user clicking a single-pixel of the material in a generated false RGB image of the captured hyperspectral image and adding a label for the class. Depending on the application and scenario, the user can add any number of semantic classes, but there is an increase in computational cost for adding classes to the spectral database, as shown in Table \ref{table: Time comp}. This eliminates the need to retrain or recalibrate any of the algorithms after a class addition.

Each label of a semantic class has a unique spectral shape and we want to find the pixels in the hyperspectral image which have a high correlation to labels in the database. Spectral similarity is the amount of correlation between two different spectra. If the similarity between a semantic class and a pixel is high, then the pixel is associated with the semantic class. We also define a tunable \emph{variance} parameter which is used to limit the allowable dissimilarity between a pixel and the reference spectra. If this \emph{variance} parameter is set high, it allows for more pixels to be classified into one of the defined semantic classes (shown in Fig. \ref{fig:Threshold value comparision}). This is the only parameter that needs to be tuned for generating the segmented image, and this can be done during run-time. In comparison, machine learning models have many parameters that must be tuned to achieve accurate results and the effect of tuning each of the multitude of parameters can be hard to predict and understand \cite{khan2020survey}.  

The framework was designed so that any spectral similarity algorithm can be used for classifying the pixels in the image. This allows the user to decide which algorithm works best for their environment and constraints. Some deciding factors can be computational cost, noise, and spectral similarity between reference spectra. The focus of this work is not on reviewing spectral similarity algorithms, so minimal analysis is done in comparing the results of the different algorithms. In the initial implementation, we used the Spectral Angle Mapper (SAM) algorithm to classify the hyperspectral image, as it is one of the most widely used spectral similarity algorithms \cite{Meneses2000SpectralCM}. 

\subsection{Semantic Image Processing}
This part of the framework was designed to be independent of the method used to generate the segmented images. This allows for the extraction of features from any semantically classified image using our pipeline. The inputs to this part of the framework are the segmented images with the associated semantic class labels. For ease of visualization, each label has an associated RGB value.

\subsubsection{Edge and Contour Detection}
\label{section: Edge and Contour Detection}
The first step in the semantic image processing pipeline is detecting edges. It is assumed that each semantic class has a unique label value in the segmented image. The edges are detected by determining the locations where there is a change in the label value. For each pixel, the surrounding 8 pixels' label values are compared to the center pixel. If all of the surrounding pixels have the same label as the center pixel, the center pixel is not considered an edge. If one of the surrounding pixels has a different label compared to the center pixel, the pixel is considered an edge. From this knowledge, a binary image is created where white pixels correspond to edge points and black pixels correspond to non-edge points. 

After the binary image is generated, the contours are extracted through a topological analysis of the binary image with a border following algorithm \cite{suzuki1985topological}. A contour represents a closed curve joining all the boundary points (along the edge) to form a polygon.

\subsubsection{Size Based Filtering}
The average height of the camera from the ground ($h$) can be determined in meters, since the hyperspectral camera uses light field technology \cite{cai2020light}. This distance is found using Cubert's Cuvis-SDK \cite{cubertsdk}. This height along with the field of view ($FOV$) is used to calculate the overall area of the image ($A$, in $m^2$), using Eq. \ref{eq:pixel2mt}. Thus, we can calculate the corresponding area of each contour so that size-based filtering can be done in $m^2$. This is more robust than filtering in terms of pixels, since scale may change between images. Further, with the availability of the height, if we have the GPS location of the camera, we can find the GPS coordinate of each pixel in the image, enabling the representation of the map in a global coordinate frame. 

\vspace{-0.7em}
\begin{equation} \label{eq:pixel2mt}
	A = (2* h * \tan (FOV/2) )^2
\end{equation}
\vspace{-1.2em}

\subsubsection{Polygon Approximation}
As we have pixel level classification of the image, the contours generated by the previous steps are very detailed and can have a lot of vertices. The high number of vertices makes the storage, processing and sharing of the semantic map computationally expensive. To overcome this challenge we approximate the contours by reducing the number of vertices. We developed and implemented Algorithm \ref{alg:polygon approximation} to perform this task as the current state of the art algorithms such as \cite{douglas1973algorithms} are unable to preserve the shape of the contour. Our algorithm was inspired by the thick edge polygon approximation performed in \cite{10.1007/978-981-10-2035-3_54} and has a similar performance in terms of reduction of vertices, shape preservation and processing time while using \textbf{15} times less memory on average and \textbf{17} times less peak memory compared to the one in \cite{10.1007/978-981-10-2035-3_54}.

The algorithm tries to find the dominant points, points which are required for shape preservation of the polygon. The input to the algorithm is the ordered pair of points of the polygon. Initially, the second point of the polygon is marked as the point of interest and the third point as the end point. Then the distance of the point of interest is calculated from a line between the initial (first point of the polygon) and the end point (Line 6). If the distance of the point of interest is less than a threshold, it is marked as non-dominant. Then the index of the end point and point of interest are increased by 1 and the new point of interest is tested for dominance (Line 7-9). When the distance of the point of interest is greater than the threshold, it is marked as dominant, and made the new initial point (Line 10-14). The above process is repeated until all the vertices are checked for dominance.

\vspace{-.25em}

\begin{algorithm}
\caption{Polygon Approximation}
\label{alg:polygon approximation}
 \hspace*{\algorithmicindent} \small \textbf{Input:} Points representing a polygon, thickness threshold \\
 \hspace*{\algorithmicindent} \small \textbf{Output:} Points representing approximated polygon
\begin{algorithmic}[1]
    \State $st_{idx} = 1$
    \State $end_{idx} = st_{idx} + 2 $
    \State $pt_{idx} = st_{idx} + 1 $
    \State $size =$ number of vertices in contour   
    \While{$start_{index} \leq size$}
        \State $dist$ = distance of $pt_{idx}$ from line [$st_{idx}, end_{idx}$]
        \If{$dist \leq threshold$}
             \State $end_{idx} \gets end_{idx} + 1$
            \State $pt_{idx} \gets pt_{idx} + 1$
        \Else
            \State Mark $pt_{idx}$ as dominant
            \State $st_{idx} = pt_{idx}$
            \State $end_{idx} = st_{idx} + 2 $
            \State $pt_{idx} = st_{idx} + 1 $   
        \EndIf
    \EndWhile
\end{algorithmic}
\end{algorithm}
\vspace{-1.5em}

\begin{figure*}
\centering
\begin{subfigure}{.5\columnwidth}
\includegraphics[width=\columnwidth]{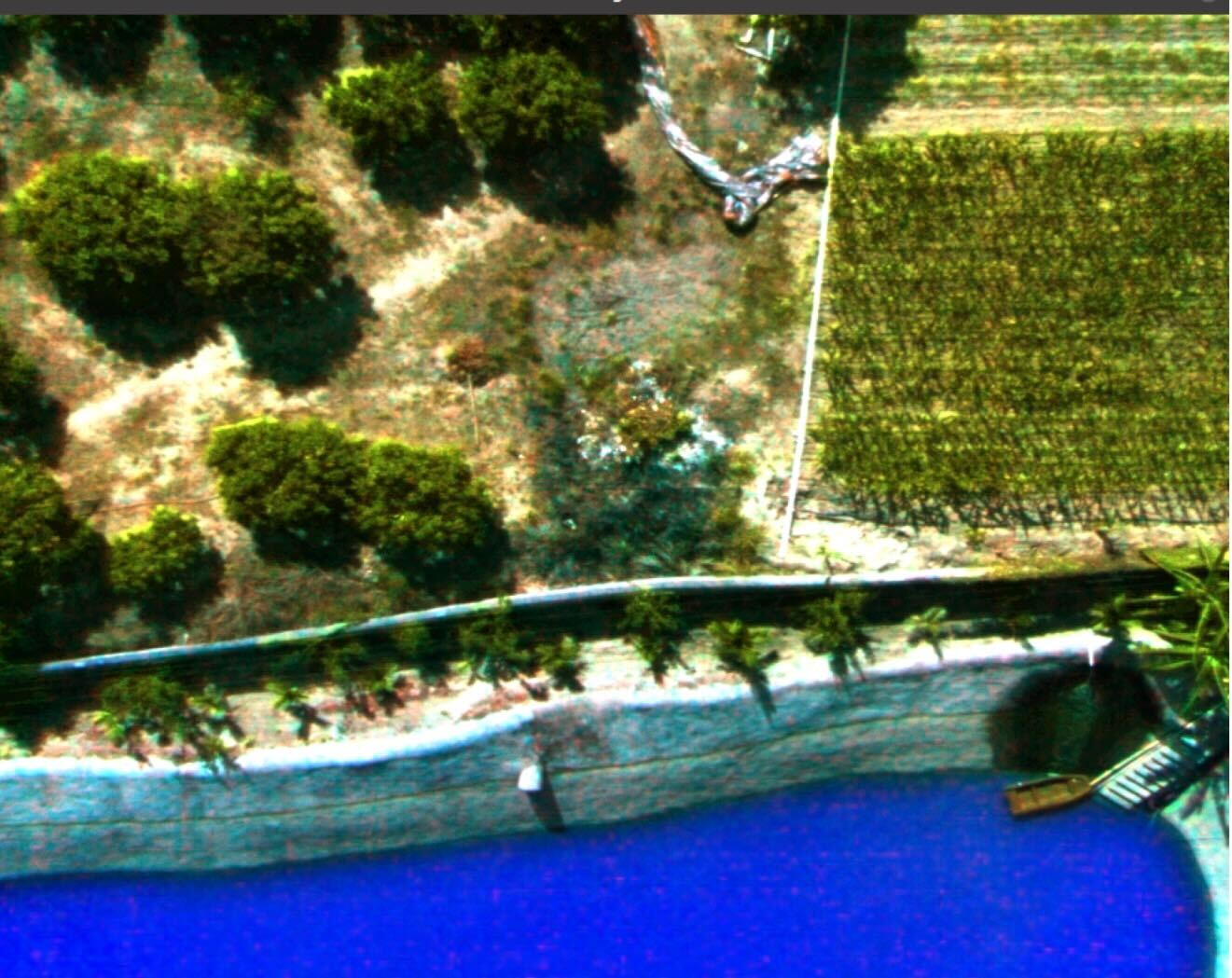}%
\caption{Standard RGB False Image}
\label{fig:RGB Image}
\end{subfigure}\hfill%
\begin{subfigure}{.5\columnwidth}
\includegraphics[width=\columnwidth]{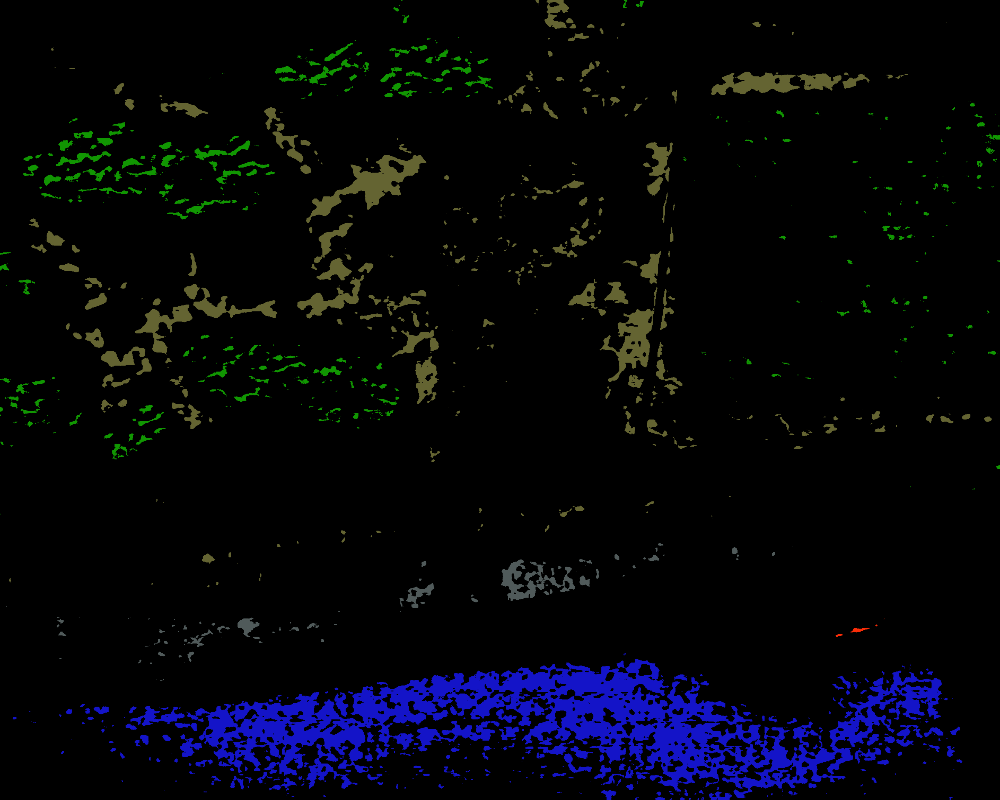}%
\caption{ \emph{Variance} parameter $= 5$}
\label{fig:thresh_5}
\end{subfigure}\hfill%
\begin{subfigure}{.5\columnwidth}
\includegraphics[width=\columnwidth]{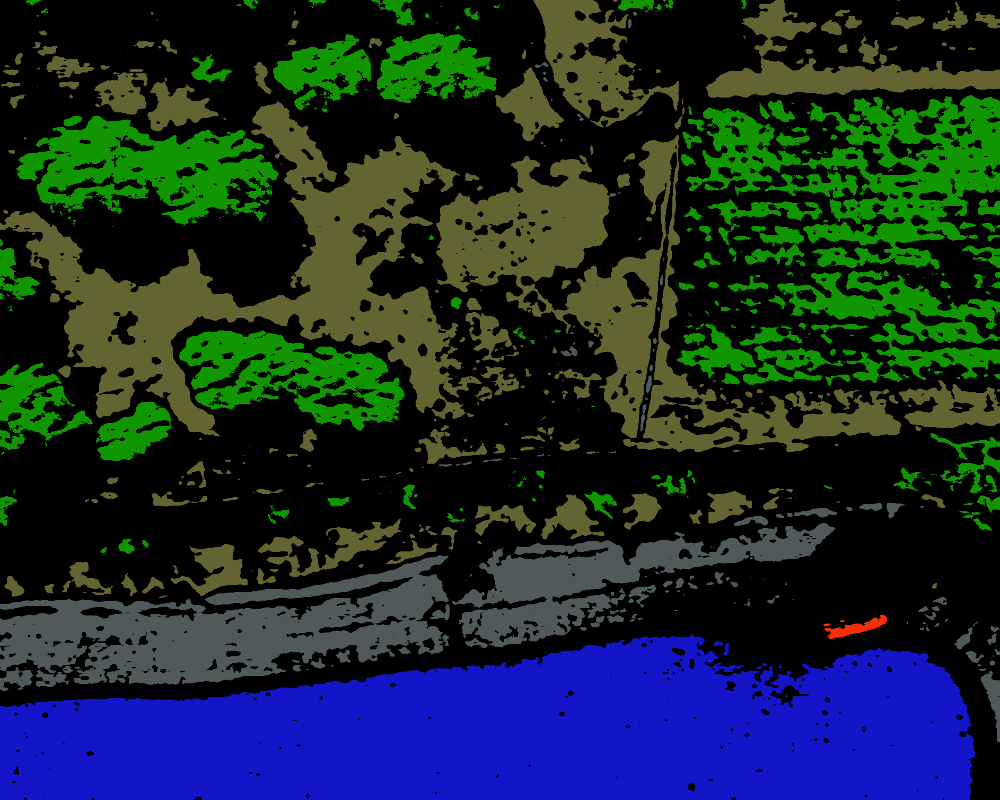}%
\caption{ \emph{Variance} parameter $= 10$}
\label{fig:thresh_10}
\end{subfigure}\hfill%
\begin{subfigure}{.5\columnwidth}
\includegraphics[width=\columnwidth]{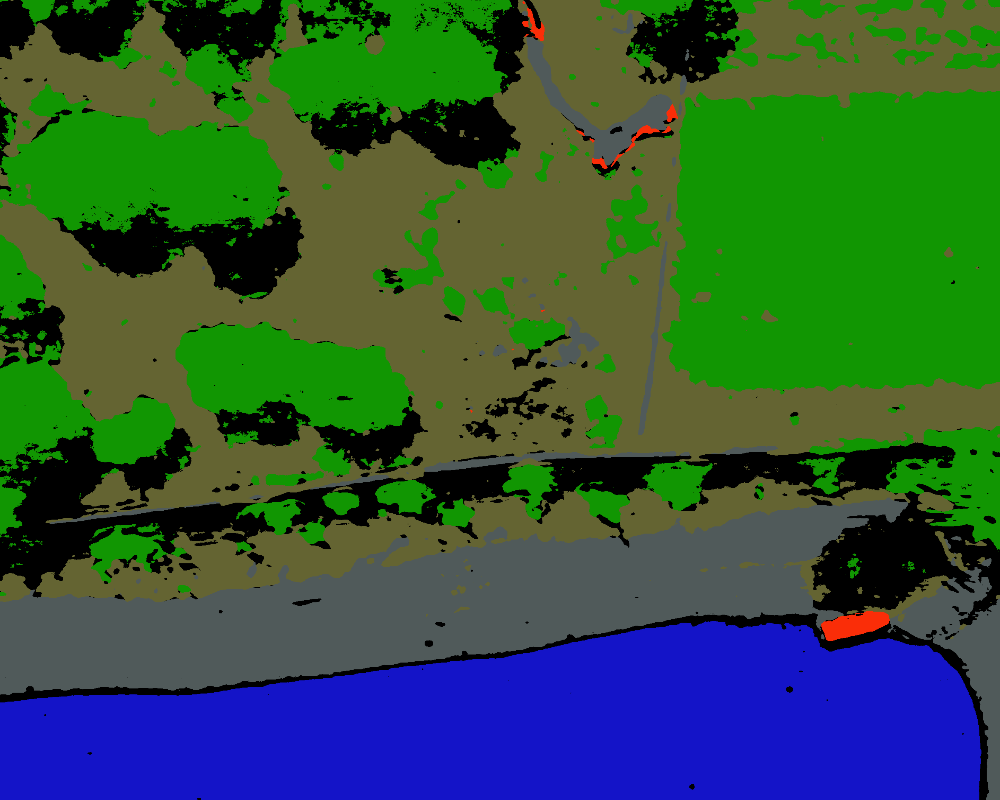}%
\caption{ \emph{Variance} parameter $= 20$}
\label{fig:thresh_20}
\end{subfigure}%
\newline
\colorbox{black}{\textcolor{black}{a}} Unknown
\colorbox{Concrete}{\textcolor{Concrete}{a}} Concrete
\colorbox{Ground}{\textcolor{Ground}{a}} Ground
\colorbox{Vegetation}{\textcolor{Vegetation}{a}} Vegetation
\colorbox{Water}{\textcolor{Water}{a}} Water
\colorbox{Wood}{\textcolor{Wood}{a}} Wood
\caption{
Impact of different \emph{variance} parameter values on hyperspectral image classification. The \emph{variance} parameter is the allowable dissimilarity between a pixel in the image and spectral shape of the reference class. It can be seen as the \emph{variance} parameter is increased, more pixels are classified, as larger dissimilarity is allowed.}
\label{fig:Threshold value comparision}
\vspace{-0.8em}
\end{figure*}

\begin{figure*}%
\centering
\begin{subfigure}{.5\columnwidth}
\includegraphics[width=\columnwidth,height = 1.35in]{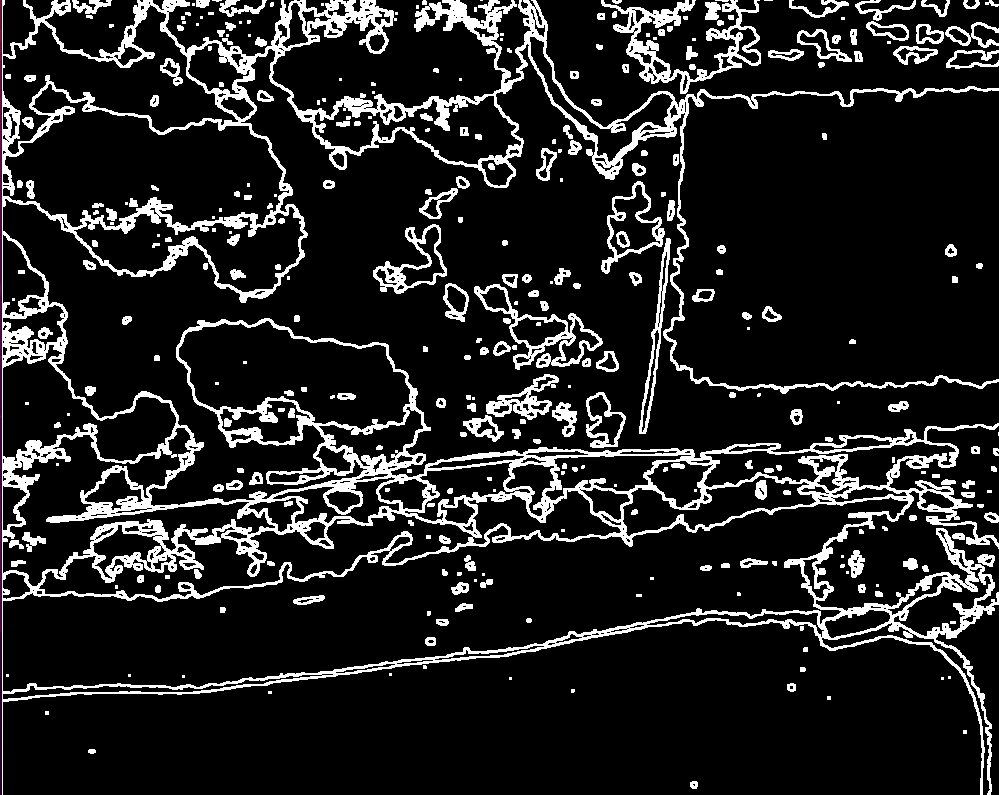}%
\caption{Binary image showing edge \newline detection on segmented image}
\label{fig:Edge Detection}
\end{subfigure}\hfill%
\begin{subfigure}{.5\columnwidth}
\includegraphics[width=\columnwidth,height = 1.35in]{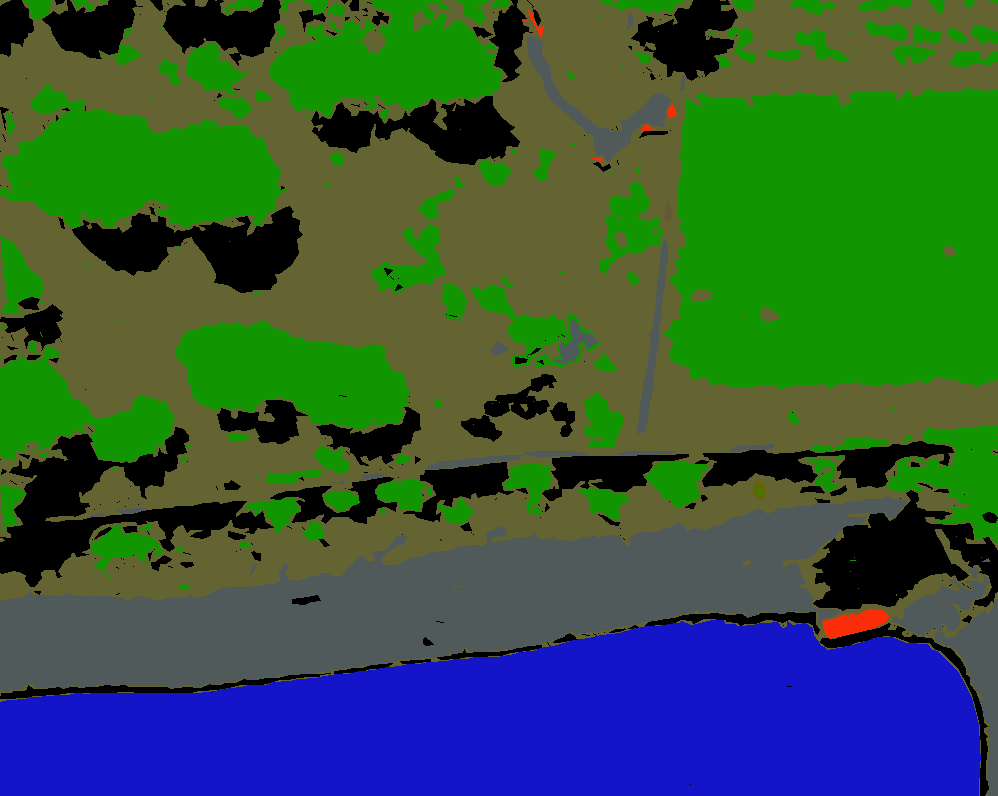}%
\caption{Extracted Contours represented as polygons }
\label{fig:contour_0}
\end{subfigure}\hfill%
\begin{subfigure}{.5\columnwidth}
\includegraphics[width=\columnwidth,height = 1.35in]{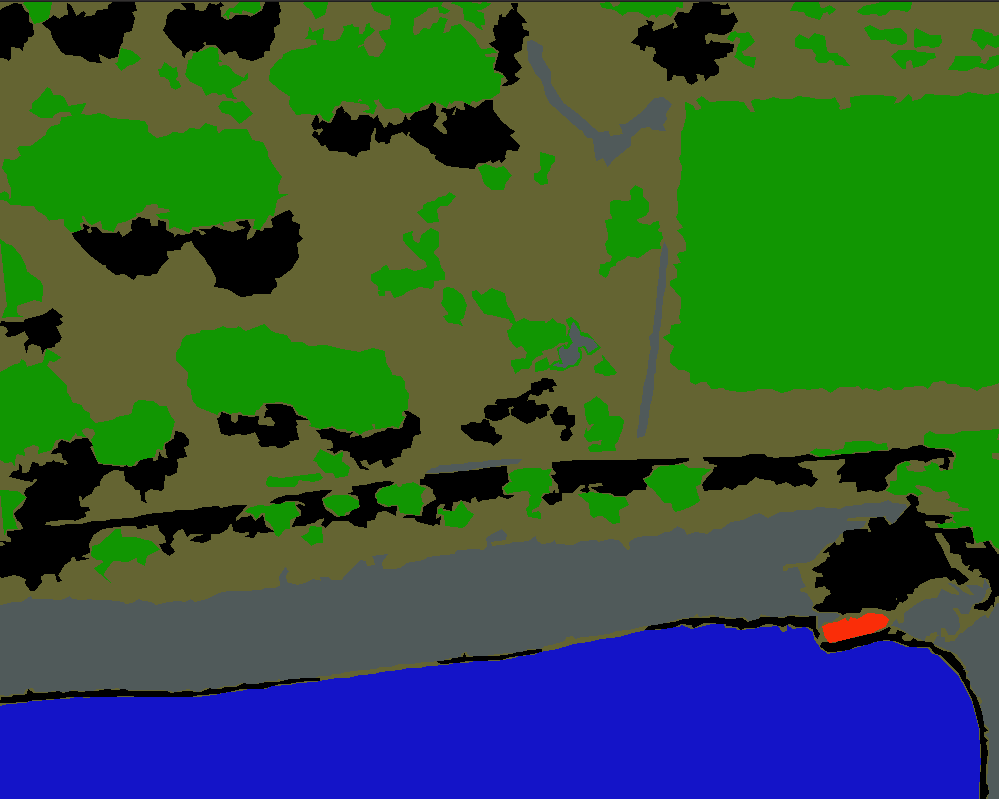}%
\caption{Polygons remaining after using an \emph{area filter} of $0.2 m^2$ }
\label{fig:contour_0.2}
\end{subfigure}\hfill%
\begin{subfigure}{.5\columnwidth}
\includegraphics[width=\columnwidth,height = 1.35in]{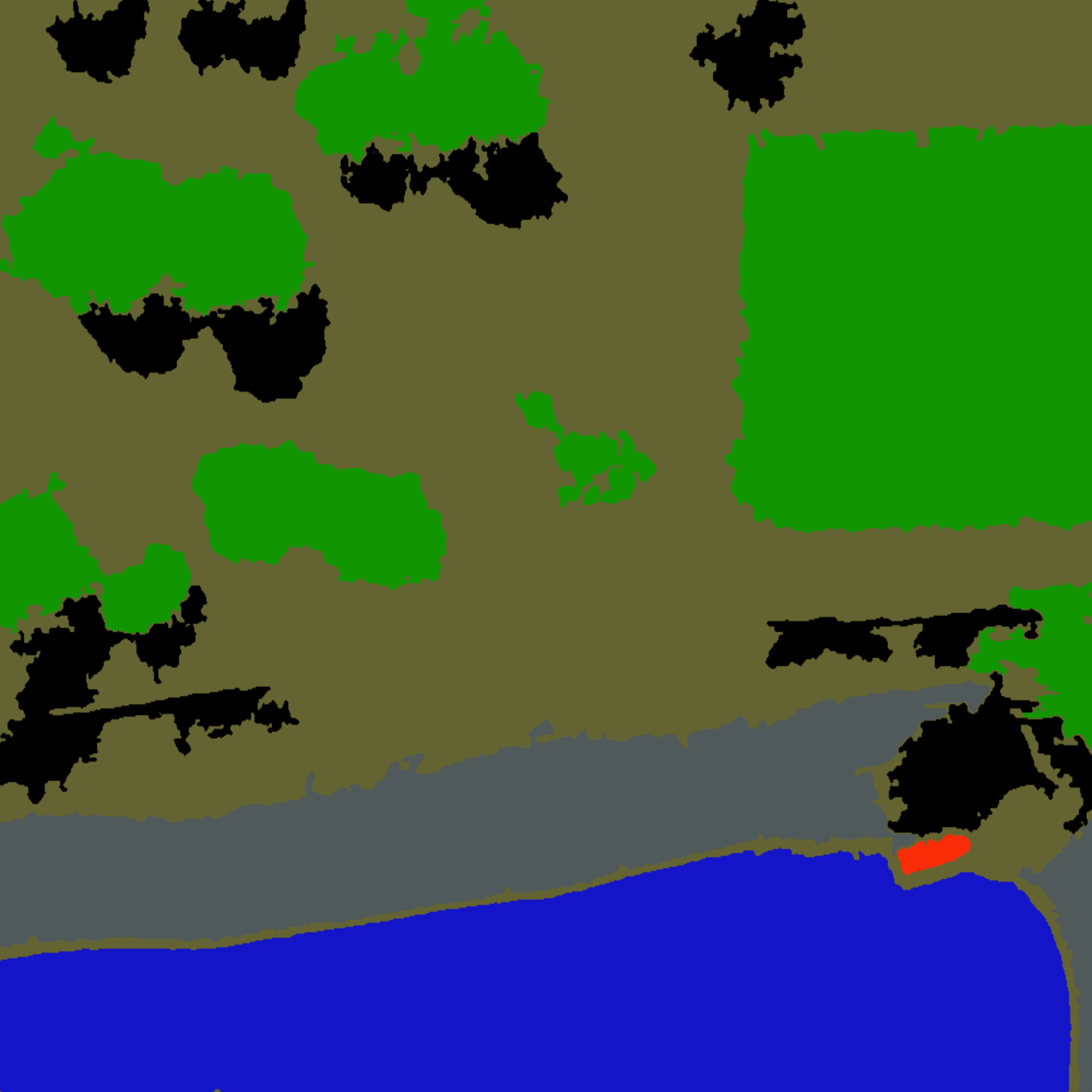}%
\caption{Polygons remaining after using an \emph{area filter} of $2.5 m^2$ }
\label{fig:contour_2.5}
\end{subfigure}%
\caption{Image processing pipeline to generate and filter the contours based on size. A small \emph{area filter} value of $0.2m^2$ can remove the noise as seen in Fig. \ref{fig:contour_0.2}. Fig. \ref{fig:contour_2.5} shows that based on the resolution required filtering can be performed for larger area contours as well.}
\label{fig:Edge Detection and Contour Extraction}
\vspace{-1.5em}
\end{figure*}

\subsection{Semantic Mapping}
The above defined pipeline outputs 2D polygons with associated semantic classes, which are used to generate a representation of the environment. Our aim is to create a general purpose semantic map which can be used to generate input for various applications like path planning, semantic inference and decision making. 

To achieve the above-mentioned requirements the polygonal features are stored in a hierarchical manner. We start with a global set of meaningful human-recognizable ``things" that are used to define the world of interest. Within that world, we identify various subsets that are of particular relevance to the domain of interest. Each of the subsets above can be considered as a ``class" in the object-oriented programming paradigm, and the subset relationships can be considered as the ``inheritance" relationships. 

Some of the basic \emph{Entities} used are described below. 
\begin{itemize}
    \item \emph{feature} - A subset of the world that represents the human recognizable elements on the ground or region that we are operating in, ex. grassy fields or rivers.
    \item \emph{instanceLabel} - They are the semantic classes that define a \emph{feature}. These labels can be obtained from the hierarchical instance label definition, with each label having unique properties about the class it represents \cite{ATLASLabelSEt}. Some examples of \emph{instanceLabels} include landscape, vegetation, obstacle, etc.  
    \item \emph{shape} - Defines the shape of the \emph{feature} to represent the geometry and geographic specification. The shape of all features are represented as polygons. 
\end{itemize}

The polygons obtained after the image segmentation are used to initialize the \emph{feature}'s in the map. The different classes which are used for segmentation provide the \emph{instanceLabel} information. The hierarchical representation of the map enables the generation of an ontology to perform inference and decision making. Further, the geometry information along with the semantic information can be used to create a traversability graph or an occupancy grid based cost maps for path planning. The focus of the paper is to show how the input to create the semantic map is generated and the applications of the semantic map are not discussed here.

\section{Experiments}

\begin{itemize}
    \item Hardware: The hyperspectral camera used was the Cubert Ultris X20 Plus. The wavelength range on the camera is 350-1000 nm and has 164 bands. The resolution of each image is 1886 x 1886 pixels and the data is captured at $4$Hz. The snapshot hyperspectral camera has a global shutter and 12 bit data, but the data is down-sampled to 8 bit resolution for this work. Images were processed on a computer with a NVIDIA GeForce RTX 3070 TI GPU and an Intel i7-12700K CPU.
    \item Software: The code for this work was written in C++ and CUDA, and was run on a Linux OS. There is a CPU-only version of the code and a version that can utilize a GPU to improve performance speed. OpenCV \cite{opencv_library} was used for various parts of our image processing framework. GTK was used to develop the graphical user interface to facilitate visualization and analysis of the hyperspectral images. The Cuvis-SDK from Cubert was used to convert hyperspectral images from the Ultris X20 Plus into a datatype supported by OpenCV. The framework was designed so that other hyperspectral cameras can also be used. Our code and a video demonstrating the capabilities of our framework can be found here: \url{https://github.com/tamu-edu-students/HyperTools}.  
\end{itemize}

The pipeline is tested on the Cornfields image set provided from Cubert. We start without any semantic classes in the reference database. The user can label one pixel from each of the relevant classes using the GUI designed. The current results are shown using the Spectral Angle Mapper algorithm and the effect of different \emph{variance} parameters is shown in Fig. \ref{fig:Threshold value comparision}. It can be seen that increasing the \emph{variance} parameter value decreases the unclassified pixels (shown in black), seen in Fig. (\ref{fig:thresh_5},\ref{fig:thresh_10},\ref{fig:thresh_20}). The user can tune the \emph{variance} parameters and choose a different spectral similarity algorithm, if required by their specific application. We decided on using a \emph{variance} parameter value of $20$ for the next steps, as it provided the best results for the particular data set. 

\begin{figure}
	\begin{subfigure}{.24\textwidth}
	\centering
    	\includegraphics[width = 01\textwidth]{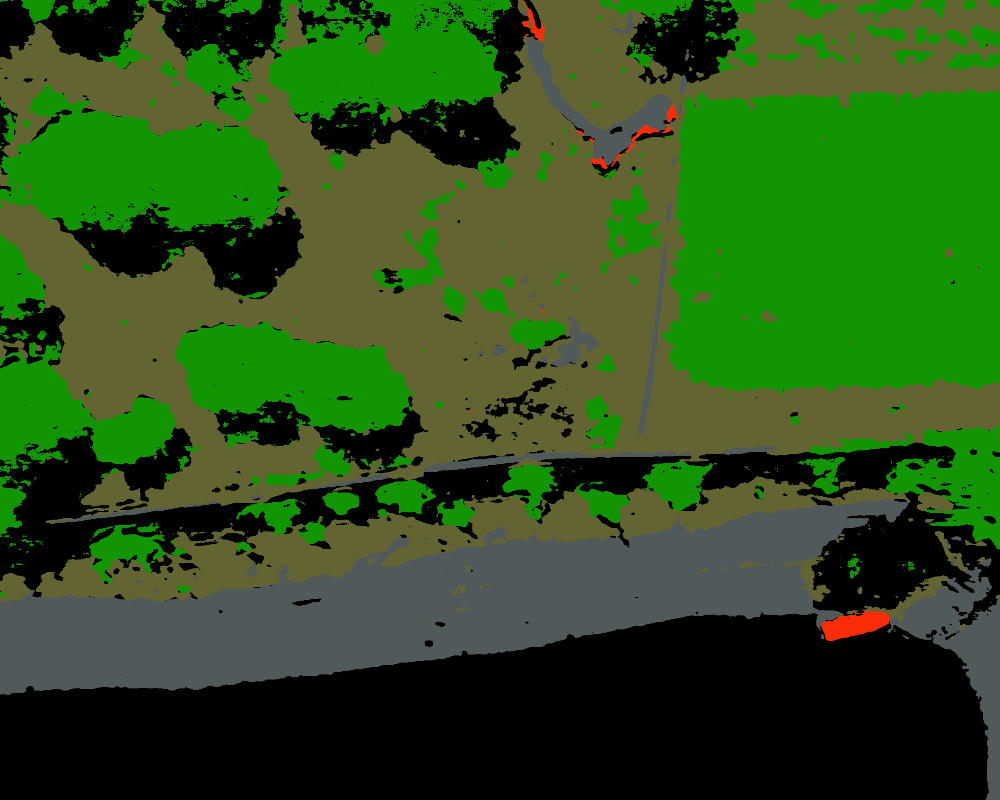}
		\caption{Before water is defined as a class in the reference database, the relevant pixels are classified as unknown. }
		\label{fig:20_no_water}
	\end{subfigure}
	\begin{subfigure}{.24\textwidth}
	\centering
	    \includegraphics[width = 01\textwidth]{images/thresh_20.png}
		\caption{After water is defined, the relevant pixels are classified as part of the corresponding class. }
		\label{fig:Water classified}
	\end{subfigure}	
	\caption{Example of segmented images, before and after adding a new Semantic Class during run-time}
    \label{fig:User adding unknown class}
    \vspace{-1.0em}
\end{figure}

Further, if the user observes unclassified pixels, they can label and add them as a new class to the reference database during run-time and re-run the classifier. This can be seen in Fig. \ref{fig:User adding unknown class}. Due to absence of water in the reference database the body of water is not classified (Fig. \ref{fig:20_no_water}), but the user can add it to the database in run-time and classify it (Fig. \ref{fig:Water classified}). 

\begin{table*}
\centering
\begin{tabular}{|P{0.12\textwidth}|P{0.16\textwidth}|P{0.13\textwidth}|P{0.1\textwidth}|P{0.09\textwidth}|P{0.12\textwidth}|P{0.1\textwidth}|}
\hline
Number of \newline Semantic Classes & Classification Algorithm (SAM) (s)  &Edge Detection (s) & Contour \newline Extraction (s) & Size \newline Filtering (s) & Polygon \newline Approximation (s) & Total Time (s)\\\hline
2 (Vegetation \& Water)  & 0.4552  & 0.0216 & 0.0042 & 0.0001 & 0.0091 & 0.4907 \\\hline
5 (All classes defined in Fig. \ref{fig:Threshold value comparision}) & 0.6978  & 0.0221 & 0.0058 & 0.0002 & 0.0224 & 0.7478\\\hline
\end{tabular}
\caption{\label{table: Time comp}Performance of each step of the proposed framework was analyzed with a different number of semantic classes. Increase in the number of classes increases the classification time. Also, classification is the most time intensive part taking around $90\%$ of the overall time. }
\vspace{-1.5em}
\end{table*}

Fig. \ref{fig:Edge Detection}, shows the binary image that is generated as an output from the edge detection algorithm. This is used as the input for the contour extraction, as this is performed using the built in OpenCV function (cv::findContours). 

\begin{figure}
\centering
\begin{subfigure}{.45\columnwidth}
\includegraphics[width=\columnwidth,height = 1.1in]{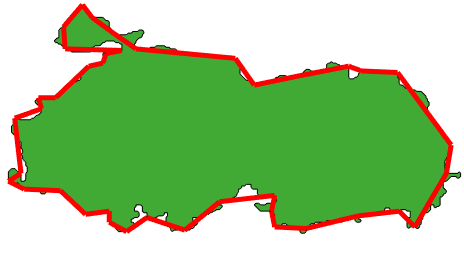}%
\vspace{1em}
\caption{\emph{thickness threshold} $= 100$,\newline Number of vertices $= 41$}
\label{fig:approx3}
\end{subfigure}\hfill%
\begin{subfigure}{.5\columnwidth}
\includegraphics[width=\columnwidth,height = 1.35in]{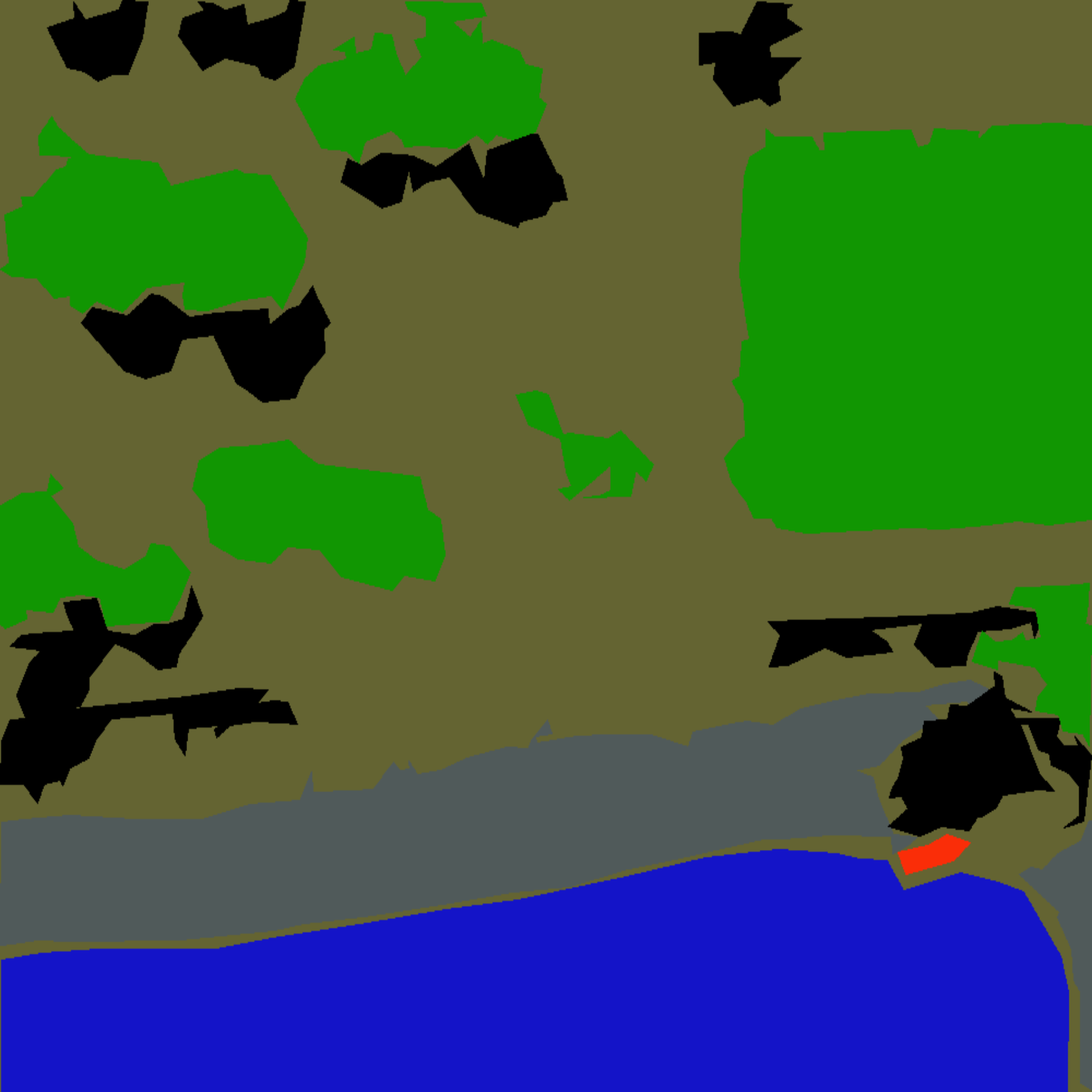}%
\caption{All filtered contours approximated using \emph{thickness threshold} $= 100$}
\label{fig:endImg}
\end{subfigure}%
\caption{Performance of the polygon approximation algorithm can be see in Fig. \ref{fig:approx3}. The number of vertices of the original polygon are reduced from $500$ to $41$, while preserving the shape. Fig. \ref{fig:endImg} shows the segmented image at the end of the image processing pipeline.}
\label{fig:polygon approximation pipeline}
\vspace{-1.2em}
\end{figure}

Fig. (\ref{fig:contour_0}, \ref{fig:contour_0.2}, \ref{fig:contour_2.5}) show the size based filtering of the extracted contours. User can define the minimum area of the contour based on the desired resolution and tolerance to noise. We can see that as the threshold area increases, we are left with fewer contours, and the area occupied by the eliminated contour is classified as the parent contour's class. 
\newpage
The contours are then passed through the polygon approximation algorithm, to reduce the number of vertices while preserving the shape. Fig. \ref{fig:approx3} shows how the algorithm can successfully reduce the number of vertices of a polygon (500 to 41 in the example), while preserving its original shape. We can increase the \emph{thickness threshold} value to further reduce the number of points in the approximated polygon or use a lower value to have a better shape approximate more number of vertices in the approximated polygon.  Fig. \ref{fig:endImg} shows image captured by the hyperspectral camera after all going through all the image processing steps. 

\begin{figure}
\includegraphics[width=\columnwidth]{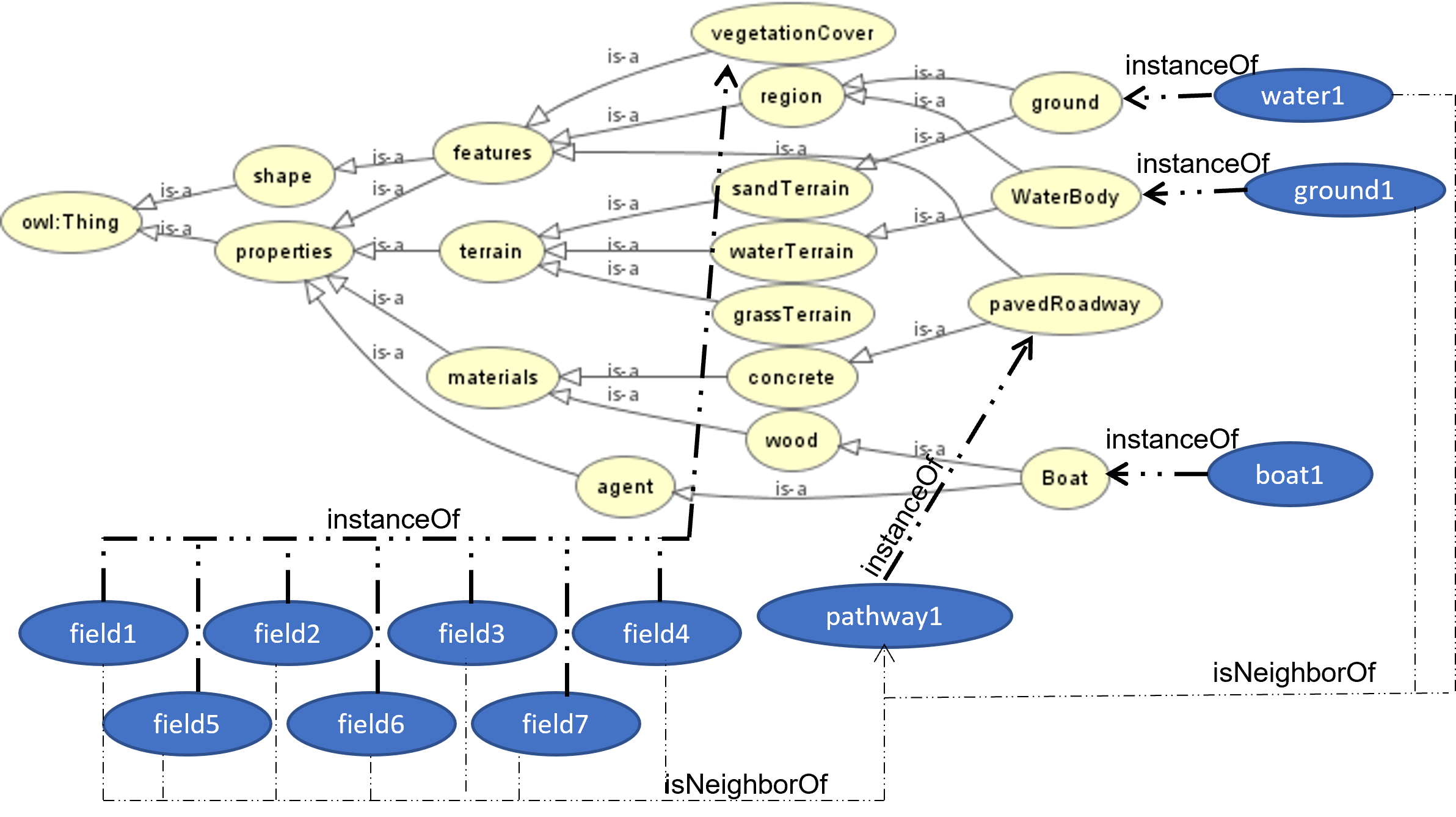}
\caption{Ontology graph generated from the semantic map. The instances are populated from the output of the semantic segmented image (Fig. \ref{fig:endImg})}
\vspace{-1.5em}
\label{fig:ontology}
\end{figure}

Fig. \ref{fig:ontology} shows a part of the ontology graph created from the semantic segmented image (Fig. \ref{fig:endImg}). The polygonal features and their semantic labels are used to instantiate the hierarchical representation of the world. We want to focus on the capability to generate inputs for a semantic map and not the use of the map itself and thus, it is not discussed in detail. 

The time taken by the separate algorithms in our pipeline is shown in Table \ref{table: Time comp}. We observe that the addition of semantic classes in the reference database increases the classification time. Also, the most computationally expensive task is image classification, still we are able to classify and generate inputs for the semantic map at a frequency of $1$Hz, allowing the real time use of the framework. 

\begin{figure}
\centering
\begin{minipage}[c]{\columnwidth}
\centering
    \includegraphics[width=\columnwidth, height = 1.35in]{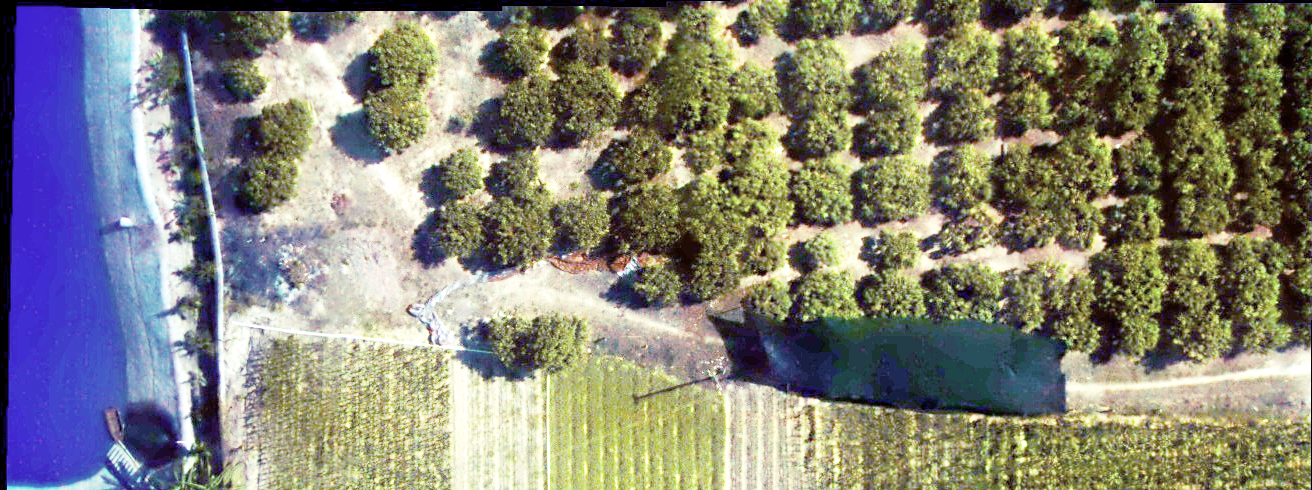}
    \caption{Multiple standard RGB images stitched together from the Cornfields data set, shown for visualization purpose only}
    \label{fig:RGB Stitched}
\end{minipage}
\begin{minipage}[c]{\columnwidth}
\centering
    \includegraphics[width=\columnwidth, height = 1.35in]{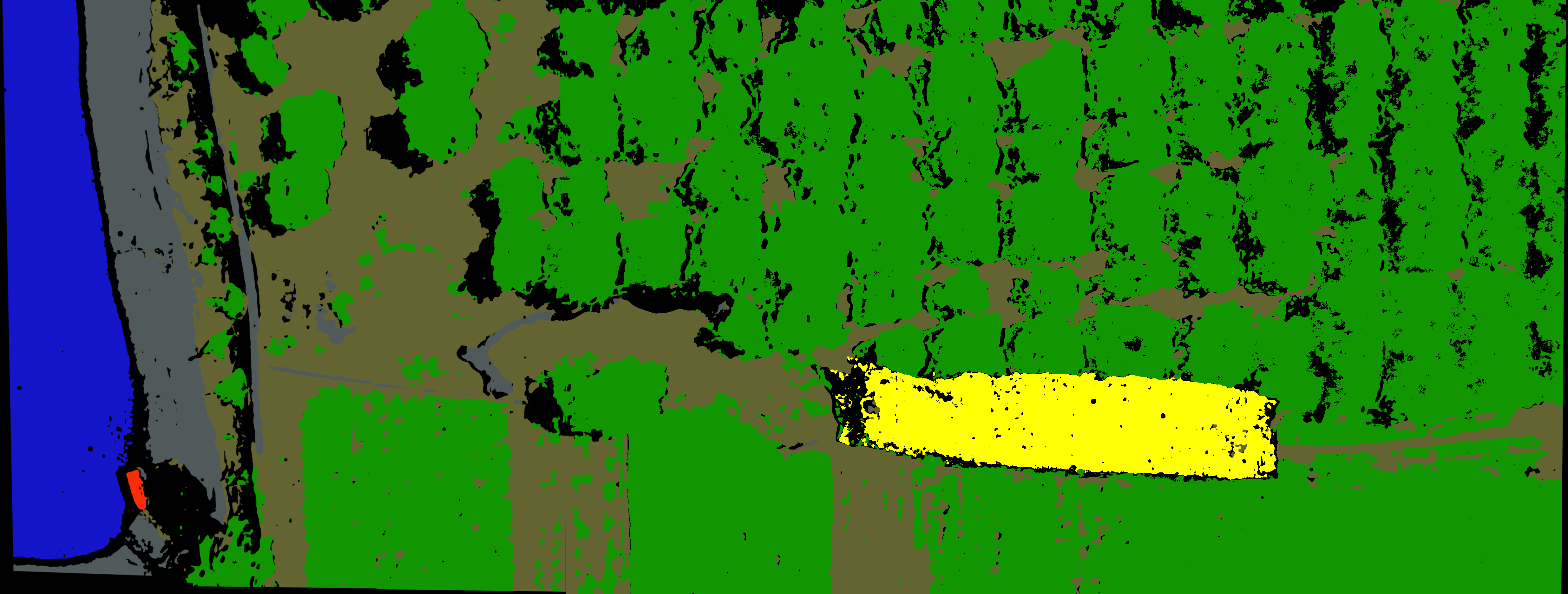}
    \newline
    \colorbox{black}{\textcolor{black}{a}} Unknown
    \colorbox{Concrete}{\textcolor{Concrete}{a}} Concrete
    \colorbox{Ground}{\textcolor{Ground}{a}} Ground
    \colorbox{Vegetation}{\textcolor{Vegetation}{a}} Vegetation
    \newline
    \colorbox{Water}{\textcolor{Water}{a}} Water
    \colorbox{Wood}{\textcolor{Wood}{a}} Wood
    \colorbox{Tarp}{\textcolor{Tarp}{a}} Tarp
    \caption{Semantic Map generated using 6 images from the Cornfields data set, using the proposed framework. Each image is 0.5GB in size, containing 27 million data points and takes about 1 second to process. Tarp was added as new semantic class in run-time as it was not observed initially. }
    \label{fig:stitch rgb}
\end{minipage}
\vspace{-1.2em}
\end{figure}

Lastly, we show the generation of the semantic map using output from the above-mentioned steps. Fig. \ref{fig:RGB Stitched} shows the aerial view of the environment for which the data was collected. Next, the sequence of aerial images captured from the hyperspectral camera were processed through the pipeline to generate the semantic map shown in the Fig. \ref{fig:stitch rgb}. 

\section{Conclusion}
We successfully demonstrate a pipeline showing how a sequence of hyperspectral images can be used to generate a semantic map of an unstructured environment. In contrast to the current state of the art, we do not rely on machine learning and show we can add new semantic classes during run-time to perform the semantic segmentation. As there is no training required, our approach does not need large, labeled training data sets. Further, we propose a pipeline to generate polygonal features from high resolution segmented images to be used as an input to the semantic map. Next, we would like to demonstrate the use of the  ontology, generated using this framework, for path planning and decision making based on the semantic properties of the various features in the environment.

Currently, due to the lack of the ground truth data for this data set, we cannot quantify the accuracy of the different classification algorithms. Thus, we would like to curate a ground truth data set for a series of hyperspectral images for testing of our algorithms and making it available to the larger research community. Lastly, we would like to compare the performance of our algorithms against the state of the art camera based semantic segmentation algorithms.




\section*{ACKNOWLEDGMENT}
This work was supported by Army Research Laboratories under contract W911NF1920243 and Bush Combat Development Complex (BCDC), Texas A\&M University. We would also like to thank Cubert for sharing the Cornfields sample data set. 

\bibliographystyle{IEEEtran}
\bibliography{IEEEabrv,references.bib}



\end{document}